\def\year{2021}\relax
\documentclass[letterpaper]{article} 
\usepackage{aaai21}  
\usepackage{times}  
\usepackage{helvet} 
\usepackage{courier}  
\usepackage[hyphens]{url}  
\usepackage{graphicx} 
\usepackage{natbib}  
\usepackage{caption} 
\usepackage{amssymb}
\usepackage{multirow}
\usepackage{amsmath}
\usepackage{subfigure}
\usepackage{array}
\def\eg{\emph{e.g.~}} 
\def\ie{\emph{i.e.~}}

\newcommand{\tabincell}[2]{\begin{tabular}{@{}#1@{}}#2\end{tabular}}

\title{Decentralised Learning from Independent Multi-Domain Labels for Person Re-Identification}
\author{
    Guile Wu, Shaogang Gong\\
}
\affiliations{
    Queen Mary University of London\\
    guile.wu@qmul.ac.uk, s.gong@qmul.ac.uk

}

\begin{document}

\maketitle

\begin{abstract}
   Deep learning has been successful for many
   computer vision tasks due to the availability of
   shared and centralised large-scale training data.
   However, increasing awareness of privacy concerns poses new challenges
   to deep learning, especially for human subject related
   recognition such as person re-identification (Re-ID).
   In this work, we solve the Re-ID problem by decentralised
   learning from non-shared private training data distributed at
   multiple user sites of independent multi-domain label spaces.
   We propose a novel paradigm called Federated Person Re-Identification
   (FedReID) to construct a generalisable global model (a central server)
   by simultaneously learning with multiple privacy-preserved local models (local clients).
   Specifically, each local client receives global model updates from the server
   and trains a local model using its local data independent from all the other clients.
   Then, the central server aggregates transferrable local model updates to
   construct a generalisable global feature embedding model
   without accessing local data so to preserve local privacy.
   This client-server collaborative learning process is iteratively
   performed under privacy control,
   enabling FedReID to realise decentralised learning without sharing distributed data nor
   collecting any centralised data.
   Extensive experiments on ten Re-ID benchmarks show that
   FedReID achieves compelling generalisation performance beyond any
   locally trained models without using shared training data,
   whilst inherently protects the privacy of each local client.
   This is uniquely advantageous over contemporary Re-ID methods.
\end{abstract}

\section{Introduction}
In recent years, deep neural network learning has achieved incredible success in many computer vision
tasks. However, it relies heavily upon the assumption that
a large volume of data can be collected from source domains and
stored on a centralised database for model training.
Despite the current significant focus on centralised data centres to facilitate big data machine
learning drawing from shared data collections,
the world is moving increasingly towards localised and private distributed data analysis
at-the-edge.
This differs inherently from the current assumption of ever-increasing availability of
centralised data and poses new challenges to deep learning,
especially for human subject related recognition
such as person re-identification~\cite{Gong2014ReID}.

\begin{figure}[t]
\begin{center}
   \includegraphics[width=0.85\linewidth]{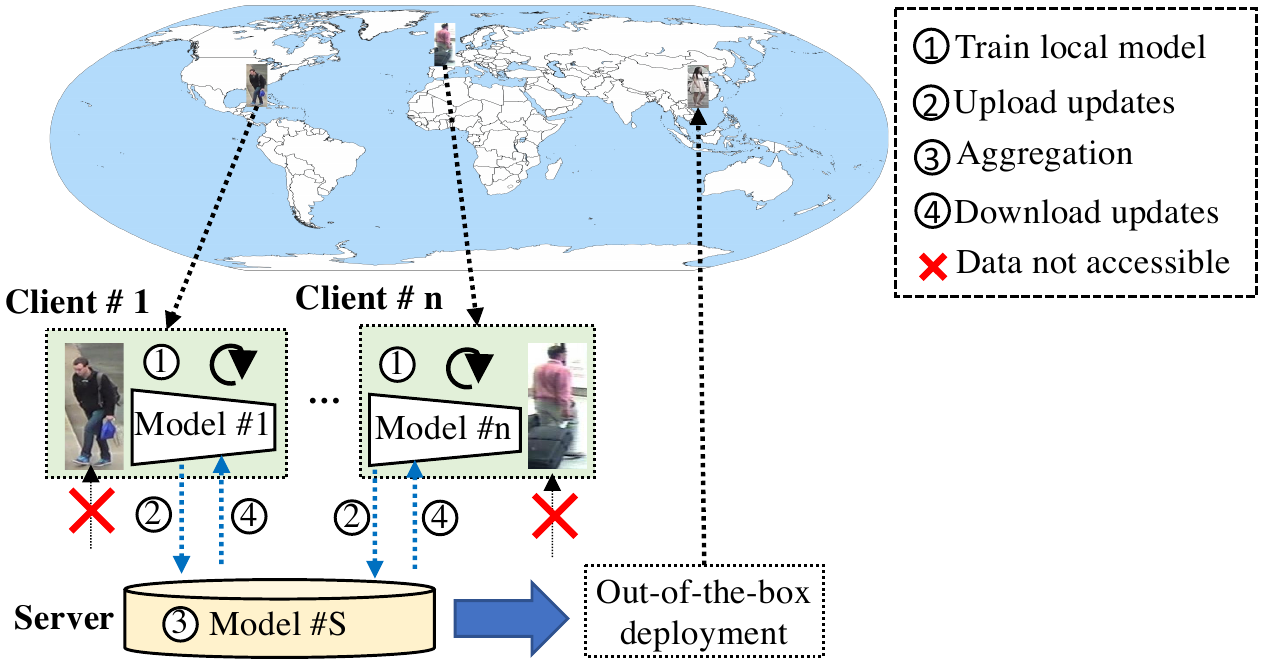}
\end{center}
   \caption{An illustration of FedReID decentralised learning.
   A client refers to a user site with private local training data,
   while a server refers to a centralised global model without any training data.
   Each client uses private local data to train a local model,
   while a server aggregates local model updates to construct a global model
   without accessing local training data.
   This client-server collaborative learning is iterative to yield
   a global feature representation for out-of-the-box deployment with privacy protection.
   }
   \label{fig:motivation}
\end{figure}

Person re-identification (Re-ID) on urban streets
at city-wide scales is useful in smart city design (e.g. population flow
management) and for public safety (e.g. find a missing person)
\cite{liu2020unity,Dong2019person,wu2020tracklet,wu2019spatio}.
Existing Re-ID methods follow either supervised learning
by collecting large-scale datasets for model training~\cite{chen2020salience,liu2020unity}
or unsupervised learning
by assembling both labelled source domain data for model
initialisation and unlabelled target domain data for model fine-tuning~\cite{wang2020unsupervised,yang2019patch}.
Although these methods have achieved promising results,
they are based on a \emph{centralised learning} paradigm,
which is inherently flawed when source datasets cannot
be shared in a centralised training protocol due to privacy protection.
This requires a new Re-ID paradigm for learning
a generalisable global model with {\em distributed collections of
non-shared data from independent multi-domain label spaces}.

In this work, we propose a fundamentally novel paradigm called Federated Person Re-Identification (FedReID)
for decentralised model learning from distributed non-sharing data of independent label spaces.
We construct a generalisable global Re-ID model (a centralised server) by
distributed collaborative learning of multiple local models (localised and
private clients) without sharing local training data nor collecting any centralised data.
As illustrated in Fig.~\ref{fig:motivation},
different cities around the world can play the roles of localised clients.
Each client receives global model updates from the central server
and trains a local model using its own set of private non-shared data.
Then, the central server aggregates local model updates
to construct a generalisable model without accessing local data.
This client-server collaborative learning process is iteratively performed,
enabling FedReID to learn a generalisable global model from decentralised data with privacy protection.
For deployment, the generalisable global Re-ID model from the server
can be deployed directly without using additional centralised data for fine-tuning.

Our {\em contributions} are:
{\bf (1)} For the first time, we introduce decentralised model
    learning from distributed non-sharing data of independent multi-domain labels for person Re-ID.
    This study can potentially benefit other computer vision tasks that require
    decentralised model learning on distributed non-sharing data with privacy protection.
{\bf (2)} We propose a new paradigm called Federated Person Re-Identification (FedReID).
    Our approach explores the principle of federated learning~\cite{konevcny2016federated},
    but is uniquely designed for decentralised Re-ID by
    reformulating the iterative client-server collaboration mechanism.
    In each local client, in addition to a local client model which consists of a
    feature embedding network for visual feature extraction and a mapping network for classification,
    we further use a local expert to regularise the training process of the local client model.
{\bf (3)} Extensive experiments on 10 Re-ID benchmarks show that
    FedReID can both protect local data privacy and achieve compelling generalisation performance,
    which is uniquely advantageous over contemporary Re-ID methods that assume
    shared centralised training data without privacy protection.

\begin{figure*}[t]
\begin{center}
   \includegraphics[width=0.99\linewidth]{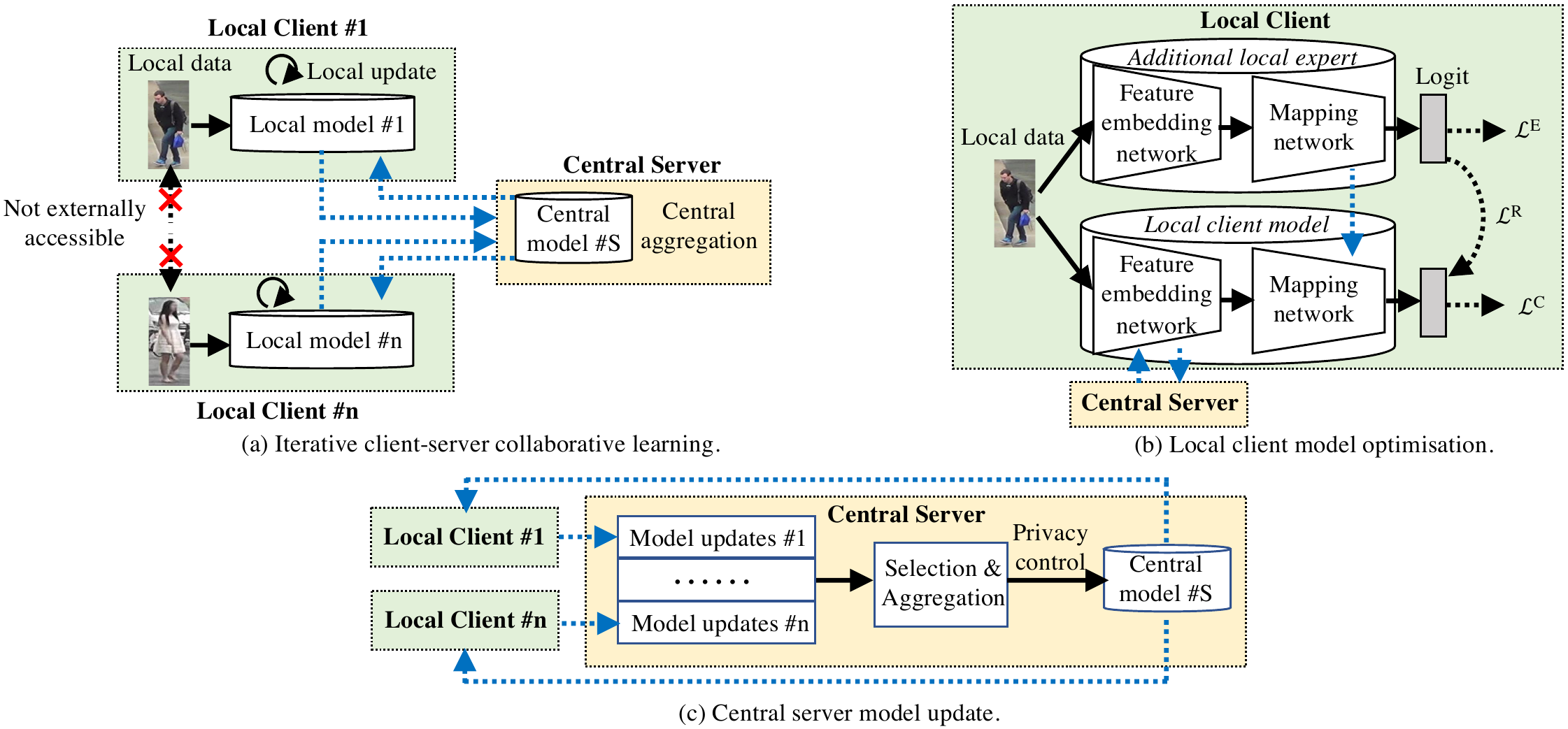}
\end{center}
\caption{An overview of the proposed Federated Person Re-Identification (FedReID).}
\label{fig:overview}
\end{figure*}

\section{Related Work}
{\noindent\bf Person Re-Identification.}
Learning generic feature representations is attractive for real-world Re-ID applications.
Conventional supervised Re-ID~\cite{chen2020salience,wu2019person}
relies heavily on centralised labelled data in target domains,
whilst cross-domain unsupervised Re-ID~\cite{wang2020unsupervised,yang2019patch}
relies on the availability of centralised labelled data from source domains for model initialisation
and unlabelled data from target domains for model fine-tuning,
so they are both impractical for out-of-the-box deployments.
More importantly, the centralised learning paradigm may not be feasible in practice when training data cannot
be shared to a centralised training process due to privacy restrictions. 
Recently, domain generalised Re-ID is proposed to learn a generic feature embedding model.
In~\cite{song2019generalizable}, \citeauthor{song2019generalizable}
follow a meta-learning pipeline to optimise a domain-invariant mapping network.
In~\cite{jin2020style}, a style normalisation module is introduced to filter out style variations.
In~\cite{xiao2016learning}, domain guided dropout is used for domain-specific knowledge selection.
However, these methods still require a centralised training process by
assembling a large pool of data from multi-domain datasets.
Different from all existing Re-ID methods, our FedReID has a
fundamentally new decentralised learning paradigm for optimising a generalised Re-ID
model through collaborative learning by communicating knowledge among
the central server model and the local client models.
Each client learns independently on distributed local private data,
while the server uses local model updates to construct a global model
without accessing local data nor collecting any centralised data,
so FedReID embraces inherently privacy protection.

{\noindent\bf Federated Learning.}
Federated learning~\cite{konevcny2016federated,mcmahan2017communication,yang2019federated,ji2019learning}
is a recently proposed machine learning technique that allows local users to collaboratively
train a centralised model without sharing local data.
Conventional federated learning aims at learning a shared model with decentralised data
for the same class label space and reducing communication cost.
For example, in~\cite{mcmahan2017communication}, \citeauthor{mcmahan2017communication} introduced Federated Stochastic
Gradient Descent (FedSGD) and Federated Average (FedAVG)
to iteratively aggregate a shared model by averaging local updates.
Our FedReID shares the merit of federated learning
but requires a fundamentally different formulation to facilitate the generalisation of
a global model for Re-ID.
In person Re-ID, each local domain is independent
(non-overlapping) from the other domains with totally different
person populations (ID space) from different locations/cities,
resulting in discrepancies in ID space and context.
Thus, we need to learn simultaneously
the non-sharing local knowledge in each local client and
the latently shared generalisable knowledge in the central server.
To this end, in FedReID, each client consists of a feature embedding network for visual feature extraction
and a mapping network for classification which is decoupled from the central aggregation process, 
while the server constructs a generalisable global feature embedding model using updates of local models.
Besides, in each local client, we additionally use a local expert to regularise the training process of the
local client model to improve the generalisation performance.

{\noindent\bf Distributed Deep Learning.}
FedReID differs significantly from distributed deep learning
\cite{mcclelland1989explorations,dean2012large,iandola2016firecaffe}.
Distributed deep learning aims at training very large scale deep networks (over billions of parameters)
using massive hardware involving tens of thousands of CPU/GPU cores
with parallel distributed computation (either model parallelism or
data parallelism), with shared large training data.
In contrast, FedReID considers the problem of optimising a
generalisable global model by asynchronous knowledge aggregation from
multi-domain locally learned models without centrally sharing training data.

{\noindent\bf Private Deep Learning.}
Private deep learning~\cite{papernot2017semi,wang2019private}
aims at constructing privacy-preserving models and preventing the model from inverse
attack~\cite{fredrikson2015model,papernot2018scalable}.
A popular solution~\cite{wang2019private}
is to use knowledge distillation to transfer private knowledge
from multiple teacher ensembles or a cumbersome teacher model to a public student model
with restricted distillation on training data.
In contrast, FedReID does not use any
centralised training data (labelled or unlabelled) for model aggregation.
Privacy is implemented intrinsically in FedReID by decentralised
model training through iterative client-server collaborative learning by
asynchronous knowledge aggregation, without central (server) data
sharing in model updates.

\section{Methodology}
{\noindent\bf Overview.}
In this work, we investigate decentralised person Re-ID,
a new problem to Re-ID which aims at optimising a generalised model via decentralised learning
from independent multi-domain label spaces without assembling local private data.
Suppose there are $N$ private datasets captured from
different locations that cannot be shared for model training due to privacy protection,
\ie there are $N$ localised clients.
As shown in Fig.~\ref{fig:overview}(a), each client updates a local model with $t_{max}$ steps separately
using its own private data and uploads the model updates to a centralised server.
The central server aggregates local model updates to construct a global model
and transmits global model updates to each client.
This client-server collaborative learning process is iteratively processed,
enabling FedReID to learn from decentralised data with privacy protection.

As shown in Fig.~\ref{fig:overview}(b), in each client,
there are a local client model and a local expert which are trained together to
improve the performance of each client.
Specifically, in the $i$-th client $(i{\in}N)$,
suppose there are $J_i$ person images of $Z_i$ identities ($Z_i \leq J_i$) in a local dataset $\mathcal X_i = \{x_{i,j}\}_{j=1}^{J_i}$,
we construct a feature embedding network $\phi(\omega_{i, t, k}^{f})$ to extract
feature representations $\mathcal V_i = \{v_{i,j}\}_{j=1}^{J_i}$ of person images:
\begin{equation}
\label{eq:client_feature}
v_{} =  \phi(\omega_{i, t, k}^{f}; x_{})
\end{equation}
where $\omega_{i, t, k}^{f}$ are model parameters the $i$-th feature embedding network
at the $t$-th local step at the $k$-th global communication epoch.
And then, we employ a mapping network $\delta(\omega_{i, t, k}^{c})$
for classification:
\begin{equation}
\label{eq:client_map}
d_{} =  \delta(\omega_{i, t, k}^{c}; v_{})
\end{equation}
where $d_{}$ is the outputted logit,
$\omega_{i, t, k}^{c}$ are model parameters of the $i$-th mapping network
at the $t$-th local step at the $k$-th global epoch.
Meanwhile,
we use an additional local expert (\{$\phi(\omega_{i, t, k}^{Ef})$, $\delta(\omega_{i, t, k}^{Ec})$\})
which is learned with local knowledge of each client 
and helps the updated local client model to learn richer knowledge
\footnote{Since the local expert will not be used for the bidirectional client-server knowledge communication,
we use ``local model'' to refer to a local client model which is used for the client-server collaboration
and use ``local expert'' to refer to the additional local expert in each client.}.
Thus, the optimisation objective $\mathcal L_i$ of the $i$-th client is formulated as:
\begin{equation}
\label{eq:client_loss}
\mathcal L_i = \mathcal L_{}^{C} + \mathcal L_{}^{E} + \mathcal L_{}^{R}
\end{equation}
where $\mathcal L_{}^{C}$ is the identity classification loss of the local client,
$\mathcal L_{}^{E}$ is the identity classification loss of the local expert,
and $\mathcal L_{}^{R}$ is the local expert regularisation from the local expert to the local client model.

As shown in Fig.~\ref{fig:overview}(c),
the central server does not use any centralised data for model optimisation.
Instead, it selects and aggregates model updates from local clients
to construct a server model $\sigma(\theta_{k})$ without accessing local private data,
where $\theta_{k}$ are model parameters of the server model at the $k$-th global epoch.

In deployment, the global feature embedding in the central server is directly used
for Re-ID matching with a generic distance metric (\eg $L2$ distance).

{\noindent\bf Client-Server Iterative Updates.}
An intuitive idea for implementing decentralised learning from multiple user sites 
is to average multiple trained local models in the parameter space to generate a global model.
However, this could lead to an arbitrarily bad model ~\cite{goodfellow2015qualitatively}.
Recent research in federated learning~\cite{konevcny2016federated,mcmahan2017communication}
shows that local client models and a central server model
can be iteratively updated for distributed model learning.
Suppose the $i$-th client is optimised using SGD
with a learning rate $\eta$, then its model parameters $\omega_{i,t+1,k}$ at the
$(t+1)$-th local step are updated as:
\begin{equation}
\label{eq:local_weight}
\omega_{i, t+1,k} = \omega_{i, t, k} - \eta \triangledown\mathcal G_{i,t+1,k}
\end{equation}
where $\triangledown\mathcal G_{i,t+1,k}$ is the set of gradient updates of the $i$-th client
at the $(t+1)$-th local step at the $k$-th global epoch.
After $t_{max}$ steps for local model updates in the clients,
the server {\em randomly selects} $S$-{\em fraction} ($S\in[0.0,1.0]$)
local clients $N_S$ (here $N_S$ is the set of selected clients)
for the server model parameters $\theta_{k}$ aggregation:
\begin{equation}
\label{eq:server_weight}
\theta_{k+1} = \frac{1}{\lceil S\cdot N\rceil}\sum_{m\in N_S} \omega_{m, t_{max}, k}
\end{equation}
where $1\leq \lceil S\cdot N\rceil \leq N$,
$\omega_{m, t_{max}, k}$ are model parameters of the $m$-th client at the $t_{max}$-th local step of
the $k$-th global epoch.
Then, in turn, each client receives $\theta_{k}$ to update the local client model:
\begin{equation}
\label{eq:local_weight_reverse}
\omega_{i,0,k+1} = \theta_{k+1}
\end{equation}
where $\omega_{i,0,k+1}$ are the model parameters of the $i$-th client
at the initial ($t$=0) step of the $k$-th global epoch.
In this way, the local clients and the server are iteratively updated for $k_{max}$
global epochs, and finally we can obtain a global model in the central server
for deployment.

{\noindent\bf FedReID Client-Server Collaboration.}
In conventional federated learning,
all model parameters in the selected client models (including feature embedding layers and
classification layers) are used to update the centralised server model (Eq.~(\ref{eq:server_weight})).
However, in decentralised Re-ID,
aggregating all model parameters might lead to performance degradation in both local and global models,
because each local dataset is usually captured in different
locations where the person ID space and context are non-overlapping.
To optimise a centralised model across different domains,
we reformulate federated learning to simultaneously consider
the non-sharing local knowledge in each client and
the latently shared generalisable knowledge in the central server.

Specifically,
we decouple $\omega_{i, t, k}^{c}$ (the mapping network) from the aggregation in Eqs.~(\ref{eq:server_weight})
and~(\ref{eq:local_weight_reverse}) 
to preserve local classification knowledge in each client,
and aggregate $\omega_{i, t, k}^{f}$ (feature embedding network) to construct a generalisable feature embedding model
for deployment.
Starting the feature embedding network of each local client model from the same initialisation,
we can accumulate updates of multiple local feature embedding networks
to find wider optima in the parameter space of a global model.
Thus, in each local client, Eq.~(\ref{eq:local_weight_reverse}) is reformulated as:
\begin{equation}
\label{eq:local_weight_moving}
\{\omega_{i,0,k+1}^{f}, \omega_{i, 0, k+1}^{c}\} =
\{\theta_{k+1}^{f} ,\omega_{i, t_{max}, k}^{c}\}
\end{equation}
Since local data in different clients are from different domains,
$\omega_{i, t, k}^{c}$ in each client corresponds to classification knowledge for different domains.
Thus, in the central server, $\omega_{i, t, k}^{c}$ does not need to be averaged.
Besides, since the feature embedding network of each local client starts from the same initialisation
(Eq.~(\ref{eq:local_weight_moving})),
accumulating local updates of feature embedding networks corresponds to aggregating
model parameters in the feature embedding space.
Therefore, the aggregation process in Eq.(\ref{eq:server_weight}) can be formulated as:
\begin{equation}
\label{eq:server_weight_feat}
\begin{split}
\theta_{k+1}^{f}
&=\theta_{k}^{f} - \frac{\eta}{\lceil S\cdot N\rceil}
\sum_{m\in N_S} \sum_{t=1}^{t_{max}} \triangledown\mathcal G_{m,t,k}^{f}\\
&=\frac{1}{\lceil S\cdot N\rceil}\sum_{m\in N_S}(\theta_{k}^{f} - 
\eta\sum_{t=1}^{t_{max}} \triangledown\mathcal G_{m,t,k}^{f})\\
&= \frac{1}{\lceil S\cdot N\rceil}\sum_{m\in N_S} \omega_{m, t_{max}, k}^{f}
\end{split}
\end{equation}
where $\theta_{k}^{f}$ are model parameters of the feature embedding network of the central server model,
$\triangledown\mathcal G_{m,t,k}^{f}$ is the set of gradient updates of the $m$-th local
feature embedding network at the $t$-th local step at the $k$-th global epoch.

{\noindent\bf Optimisation Objective.}
In FedReID, the central server does not use any centralised data for model fine-tuning,
so its optimisation process is the selection and aggregation process as formulated in Eq.~(\ref{eq:server_weight_feat}).
In each local client, as shown in Fig.~\ref{fig:overview}(b) and Eq.~(\ref{eq:local_weight_moving}),
the local client model receives global model updates from a central model.
Then, we use a cross-entropy loss to learn classification knowledge:
\begin{equation}
\label{eq:loss_id}
\mathcal L_{}^{C} = -\sum_{z=1}^{Z_i}{y_{z}}log\frac{exp(d_{z})}{\sum_{b=1}^{Z_i}{exp(d_{b})}}
\end{equation}
where $y_{z}$ is the ground-truth label and $d_{z}$ is the logit over a class $z$.
To further improve generalisation of the local client model,
we use a local expert to help the local client model to learn richer knowledge via
knowledge distillation~\cite{hinton2015distilling},
which also potentially facilitates the aggregation in the global model.
Specifically, the local expert with model parameters $\{\omega_{i,0,k+1}^{Ef}, \omega_{i, 0, k+1}^{Ec}\}$
is initialised with Eq.~(\ref{eq:local_weight_expert})
and also optimised with a cross-entropy loss with Eq.~(\ref{eq:loss_id_expert}):
\begin{equation}
\label{eq:local_weight_expert}
\{\omega_{i,0,k+1}^{Ef}, \omega_{i, 0, k+1}^{Ec}\} =
\{\omega_{i,t_{max},k}^{f} ,\omega_{i, t_{max}, k}^{c}\}
\end{equation}
\begin{equation}
\label{eq:loss_id_expert}
\mathcal L_{}^{E} = -\sum_{z=1}^{Z_i}{y_{z}}log\frac{exp(d_{z}')}{\sum_{b=1}^{Z_i}{exp(d_{b}')}}
\end{equation}
where $d_{}'$ is the logit computed by Eqs.~(\ref{eq:client_feature}) and~(\ref{eq:client_map})
with model parameters of the local expert.
As shown in Fig.~\ref{fig:overview}(b), Eqs.~(\ref{eq:local_weight_moving}) and~(\ref{eq:local_weight_expert}),
the difference between the local client model and the local expert is that
the former receives a global model for starting feature embedding network with the same initialisation
among clients,
while the latter utilises the latest local feature embedding network
as an initialisation.
Since the local expert is only learned with local data without receiving external updates,
it performs better than the global model on the local domain but worse than the global model
on the other domains (likely overfit locally).
Thus, we use it as a regularisation to help the updated local client model to learn
richer knowledge on a specific local domain.
We feed the same batch to the two models but with separately random data augmentation
and compute soft probability distributions for the local client model
($\mathcal P_{}$) and the local expert ($\mathcal Q_{}$) as:
\begin{equation}
\label{eq:soft_probability}
\mathcal P_{z} = \frac{exp(d_{z}/T)}{\sum_{b=1}^{Z_i}{exp(d_{b}/T)}},
\mathcal Q_{z} = \frac{exp(d_{z}'/T)}{\sum_{b=1}^{Z_i}{exp(d_{b}'/T)}}
\end{equation}
where $T$ is a temperature~\cite{hinton2015distilling}.
The local expert regularisation $\mathcal L_{}^{R}$ is therefore defined as the KL-divergence
between $\mathcal P_{}$ and $\mathcal Q_{}$:
\begin{equation}
\label{eq:regularisation_kl}
\mathcal L_{}^{R} = T^2\sum_{z=1}^{Z_i}\mathcal Q_{z}{\cdot}log\frac{\mathcal Q_{z}}{\mathcal P_{z}}
\end{equation}
In summary, each local client is optimised with Eq.~(\ref{eq:client_loss}) using local private data
independent from the other clients.

{\noindent\bf Privacy Protection.}
In FedReID, local sensitive data are inherently protected by
the decentralised learning process.
To further protect sensitive data from inverse attack~\cite{fredrikson2015model},
we use the white noise~\cite{geyer2017differentially} to hide the
contributions of each client in Eq.~(\ref{eq:server_weight_feat}):
\begin{equation}
\label{eq:server_weight_noise}
\theta_{k+1}^{f} = \frac{1}{\lceil S{\cdot}N\rceil}\sum_{m\in N_S} \omega_{m, t_{max}, k}^{f} + \beta\mathcal N(0, 1)
\end{equation}
where $\mathcal N(0, 1)$ is the white noise matrices with mean $0$ and variance $1$.
$\beta\in[0,1]$ is a scale factor to control 
the balance between privacy-preserving and Re-ID accuracy.
When $\beta$=$0$, the white noise is removed from the aggregation.
Moreover, in the client-server collaboration, we can also hide the collaboration information in
Eq.~(\ref{eq:local_weight_moving}) as:
\begin{equation}
\label{eq:local_weight_moving_privacy}
\{\omega_{i,0,k+1}^{f}, \omega_{i, 0, k+1}^{c}\} =
\{\theta_{k+1}^{f} + \beta\mathcal N(0, 1), \omega_{i, t_{max}, k}^{c}\}
\end{equation}

\section{Experiments}
{\noindent\bf{Datasets.}}
We used four large-scale Re-ID datasets (DukeMTMC-ReID~\cite{zheng2017unlabeled},
Market-1501~\cite{zheng2015scalable},
CUHK03~\cite{li2014deepreid,zhong2017re} and
MSMT17~\cite{wei2018person}) as non-shared local datasets in four client sites.
Each of the four local clients did not share its training data with other clients nor the server.
This is significantly different from existing domain generalised Re-ID,
where FedReID is trained with decentralised data, while existing methods are trained with centralised data.
The FedReID model was then evaluated on five smaller Re-ID datasets
(VIPeR~\cite{gray2008viewpoint}, iLIDS~\cite{zheng2009associating}, 3DPeS~\cite{baltieri20113dpes},
CAVIAR~\cite{cheng2011custom} and
GRID~\cite{loy2013person}), plus a large-scale Re-ID dataset
(CUHK-SYSU person search ~\cite{xiao2017joint}) as new unseen target domains for
out-of-the-box deployment tests.
For CUHK-SYSU, we used ground-truth person bounding box annotations for Re-ID test (not person search),
of which there are 2900 query persons and each person contains at least one image in the gallery
(both query and gallery sets are fixed removing distractors in the variable gallery sets).
On smaller Re-ID datasets, we did random half splits to generate 10 training/testing splits.
In each split, we randomly selected one image of each test identity as the query while the others as the gallery
for evaluation.
The dataset statistics are summarised in Table~\ref{table:benchmarks}.
Besides, we used CIFAR-10~\cite{krizhevsky2009learning}
for federated formulation generalisation analysis on image classification.

\begin{table}[t]
\begin{center}
\small
\begin{tabular}{c|c|c|c|c|c}
\hline
Types &Datasets & Tr. id & Tr. img& Te. id & Te. img \\
\hline\hline
\multicolumn{1}{c|}{\multirow{4}{*}{\tabincell{c}{Local\\Client\\Train}}}  & Duke & 702 & 16522 & -& -\\
& Market & 751 & 12936 & -& -\\
& CUHK03 & 767 & 7365 & - & -\\
& MSMT17 & 1041 & 30248 & - & -\\
\hline
\multicolumn{1}{c|}{\multirow{7}{*}{\tabincell{c}{New\\Domain\\Test}}}
& VIPeR & - & - & 316 & 632\\
& iLIDS & - & - & 60 & 120\\
& 3DPeS & - & - & 96 & 192\\
& CAVIAR & - & - & 36 & 72\\
& GRID & - &- & 125 & 1025\\
& Cuhk-Sysu & - & - & 2900 & 8347\\
\hline
\end{tabular}
\end{center}
\caption{The Re-ID dataset statistics.
'Tr.': Train;
'Te.': Test;
'id': number of identities;
'img': number of images.
}
\label{table:benchmarks}
\end{table}

{\noindent\bf{Evaluation Metrics.}}
We used Rank-1 (R1) accuracy and mean Average Precision (mAP)
for Re-ID performance evaluation.
Note that FedReID is designed to learn a generalised model from distributed datasets with privacy protection.

{\noindent\bf{Implementation Details.}}
In our design, the feature embedding network
is ResNet-50~\cite{he2016deep} (pretrained on ImageNet),
while the mapping network consists of two fully connected layers,
in which the first fully connected layer follows by a batch normalization layer,
a ReLU layer and Dropout.
In practice, both global and local models used a multi-head architecture in which each mapping network in each head was corresponding to one client.
By default, we set the number of local clients $N$=$4$,
client fraction $S$=$1.0$ ,
and privacy control parameter $\beta$=$0$.
These hyperparameters can be determined by different application requirements.
We empirically set batch size to $32$, maximum global communication epochs $k_{max}$=$100$,
maximum local steps $t_{max}$=$1$, and temperature $T$=$3$.
We used SGD as the optimiser with Nesterov momentum $0.9$ and weight decay 5$e$-4.
The learning rates were set to $0.01$ for embedding networks and $0.1$ for mapping networks,
which decayed by $0.1$ every $40$ epochs.
Our models were implemented with Python(3.6) and PyTorch(0.4), and trained on TESLA V100 GPU (32GB).

{\subsection{Federated Formulation Generalisation Analysis}}
To analyse the generalisation of our FedReID,
we compared FedReID with FedSGD~\cite{mcmahan2017communication}, FedAVG~\cite{mcmahan2017communication},
and FedATT~\cite{ji2019learning}.
When adapting FedSGD, FedAVG and FedATT for Re-ID,
we set the last classification layer with the maximal identity number among local datasets.
When adapting FedReID for CIFAR-10, we aggregate all layers (including classifiers, \ie using Eqs.~(\ref{eq:server_weight})
and~(\ref{eq:local_weight_reverse})) in the client-server collaboration because all local data are from the same domain.
On CIFAR-10, we employed ResNet-32 for experiments and set $t_{max}$=5.
As shown in Table~\ref{table:exp_federated},
FedReID achieves better generalisation than the other federated variants.
When training and testing on the same domain for image classification (CIFAR-10),
FedReID performs closely to the centralised joint-training, but FedReID only accesses to
local model updates with privacy protection whilst joint-training assembles together
all local data without privacy concerns.
When testing out-of-the-box on unseen new domains for decentralised Re-ID
(VIPeR and iLIDS),
FedReID performs even better than joint-training, which can be attributed to:
(1) FedReID accumulate multiple local model updates
to find wider optima in the parameter space of a global model;
(2) With the client-server collaboration, FedReID learns softer distribution knowledge.
Therefore, FedReID might get better generalisation for an unseen new Re-ID domain
while joint-training is the upper bound for testing on a source client.

\begin{table}[t]
\begin{center}
\small
\begin{tabular}{c|c|c|c}
\hline
Methods&CIFAR-10&VIPeR&iLIDS\\

\hline\hline
FedSGD & 90.72$\pm$0.04&41.0 & 65.2\\

FedAVG ($S$=0.5)& 93.04$\pm$0.19& 41.5& 65.3\\

FedAVG ($S$=1.0)& 93.21$\pm$0.08& 41.0& 65.2\\

FedATT ($S$=0.5)& 92.86$\pm$0.12& 38.3& 65.3\\

FedATT ($S$=1.0)& 92.97$\pm$0.21& 40.2& 61.3\\

FedReID ($S$=0.5)& 93.14$\pm$0.14&  45.3& 68.3\\

FedReID ($S$=1.0)& {\bf 93.31$\pm$0.11}& {\bf 46.2}& {\bf 69.7}\\
\hline

Baseline (Centralised joint)& 93.58$\pm$0.14& 44.6& 65.5\\
\hline
\end{tabular}
\end{center}
\caption{Evaluating generalisation of federated formulations.
Top-1/R1 accuracies are reported.
FedSGD means setting $S$=1.0 and $t_{max}$=1.
}
\label{table:exp_federated}
\end{table}

\begin{figure}[t]
\begin{center}
   \includegraphics[width=1.00\linewidth]{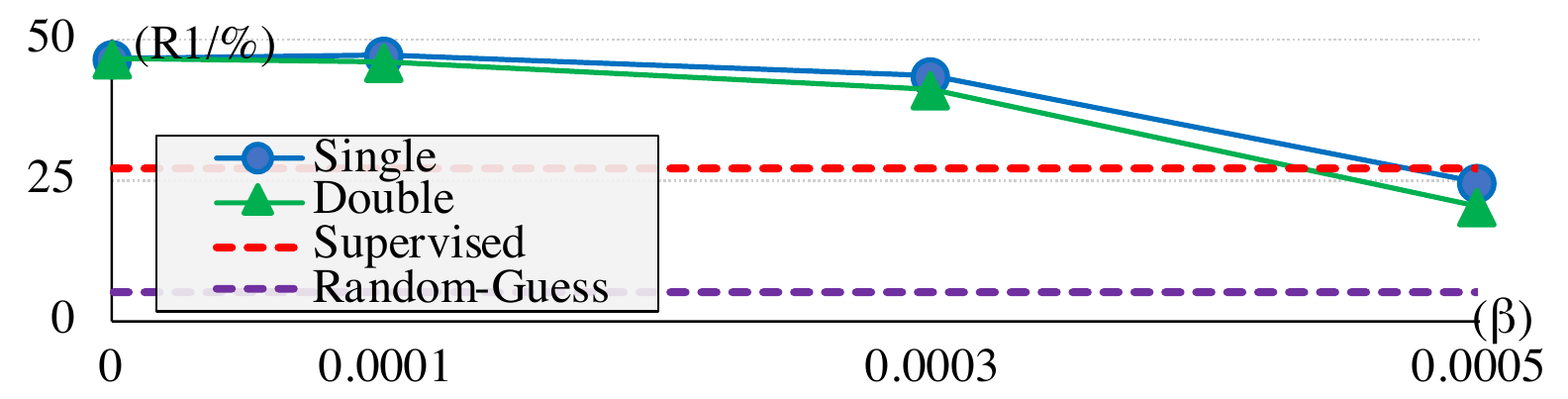}
\end{center}
   \caption{Evaluating the privacy control parameter $\beta$ on VIPeR.
   `Single': use $\beta$ in Eq.~(\ref{eq:server_weight_noise}),
   `Double': use $\beta$ in both Eqs.~(\ref{eq:server_weight_noise})
   and~(\ref{eq:local_weight_moving_privacy}),
   `Random-Guess': initialise the model with ImageNet pretrained parameters without training.
   }
   \label{fig:privacy}
\end{figure}

{\subsection{Privacy Protection Analysis}}
The inherent defensive ability of FedReID is given by decentralised learning and model aggregation.
Besides, $\beta$ in Eqs.~(\ref{eq:server_weight_noise}) and~(\ref{eq:local_weight_moving_privacy})
can further control privacy protection.
From Fig.~\ref{fig:privacy}, we can see that:
(1) R1 accuracy of FedReID gradually decreases when $\beta$ increases,
but FedReID achieves significantly better accuracy than the Random-Guess;
(2) Single $\beta$ protection performs slightly better than
double protection, but the double one protects more information;
(3) When $\beta$=$0.0005$, R1 accuracy of FedReID remains close to the overfitting local supervised
method, which indicates the compromise of accuracy and privacy.

\begin{table}[t]
\begin{center}
\small
\begin{tabular}{c|c|c|c}
\hline
\multicolumn{1}{c|}{\multirow{1}{*}{Settings}} & \multicolumn{1}{c|}{\multirow{1}{*}{Methods}}
& VIPeR & iLIDS\\

\hline\hline
\multicolumn{1}{c|}{\multirow{4}{*}{\tabincell{c}{Individuals}}}
&Client ({Duke}) & 25.0 & 56.2\\

&Client ({Market}) & 26.1 & 48.0\\

&Client ({CUHK03}) & 21.6 & 41.0\\

&Client ({MSMT})& 27.3 & 60.5\\

\hline
\multicolumn{1}{c|}{\multirow{2}{*}{\tabincell{c}{Ensembles}}} & Feat-Concatenation & 29.4 & 56.2\\

& Parameter-Average & 19.9 & 41.8\\
\hline
Decentralised & FedReID & {\bf 46.2}& {\bf 69.7}\\

\hline
Centralised & Baseline (Joint) & 44.6 & 65.5 \\
\hline
\end{tabular}
\end{center}
\caption{Comparison with individual clients and ensembles.
R1 accuracies are reported.
}
\label{table:exp_ensemble}
\end{table}

{\subsection{Comparison with Individuals and Ensembles}}
We separately trained the baseline model on four localised datasets as the individuals
and used feature concatenation and model parameter average as the ensembles.
As shown in Table~\ref{table:exp_ensemble}:
(1) FedReID significantly outperforms the
individuals and the ensembles,
which shows that the collaboration between the localised clients and the centralised
server facilitates holistic optimisation, enabling FedReID to construct a better generalisable model
with privacy protection;
(2) Compared with the centralised joint-training baseline,
FedReID achieves competitive R1 accuracies, demonstrating its effectiveness;
(3) Averaging multiple trained local models in the parameter space
leads to an arbitrarily bad model.

\begin{table}[t]
\begin{center}
\small
\begin{tabular}{c|p{0.15\columnwidth}|p{0.05\columnwidth}p{0.05\columnwidth}p{0.05\columnwidth}p{0.05\columnwidth}p{0.05\columnwidth}}
\hline
Settings &\centering{Methods} &\centering{VI.} &\centering{iL.} &\centering{3D.} &\centering{CA.} &GR. \\
\hline\hline

\multicolumn{1}{c|}{\multirow{4}{*}{\tabincell{c}{w/o privacy\\(Cross-domain\\fine-tune)}}}

&TJAIDL~ & 38.5 & - & - & - & -\\

&DSTML  &8.6 & 33.4 &  32.5 & 28.2 & -\\

&UMDL & 31.5 & 49.3 & - & 41.6 & - \\

&PAUL & 45.2 & - & - & - & -\\

\hline
\multicolumn{1}{c|}{\multirow{5}{*}{\tabincell{c}{w/o privacy\\(Centralised\\generalised)}}}

&SyRI & 43.0 &  56.5 & - & - & - \\

&SSDAL & 43.5 & -  & - & - & 22.4 \\

&MLDG$^{\dagger}$ & 23.5 & 53.8 & - & - & 15.8 \\

&CrossGrad$^{\dagger}$  & 20.9 & 49.7 & - & -  & 9.0 \\

&DIMN & 51.2 & 70.2 & - & - & 29.3\\

\hline
\multirow{2}{*}{\tabincell{c}{{\bf{w/ privacy}}\\(Decentralised)}}
&\multirow{2}{*}{FedReID}
&  \multirow{2}{*}{46.2} & \multirow{2}{*}{69.7}
& \multirow{2}{*}{67.0}  & \multirow{2}{*}{45.6}
& \multirow{2}{*}{24.2} \\
&&&&&&\\
\hline
\end{tabular}
\end{center}
\caption{Generalised Re-ID performance evaluation on VIPeR, iLIDS,
3DPeS, CAVIAR and GRID.
R1 accuracies are reported.
$^{\dagger}$: Re-ID domain generalisation results.
}
\label{table:state-of-the-art_small}
\end{table}

\begin{figure*}[t]
\centering
\subfigure[Client number]{
\includegraphics[width=0.21\linewidth]{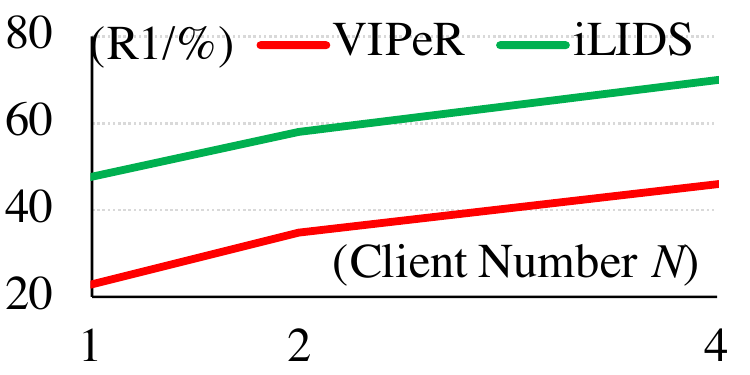}
\label{fig:client_num}
}
\subfigure[Client fraction]{
\includegraphics[width=0.21\linewidth]{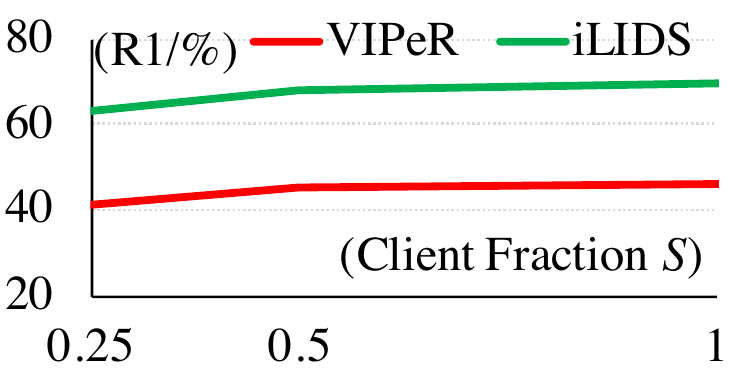}
\label{fig:client_frac}
}
\subfigure[Client local step]{
\includegraphics[width=0.21\linewidth]{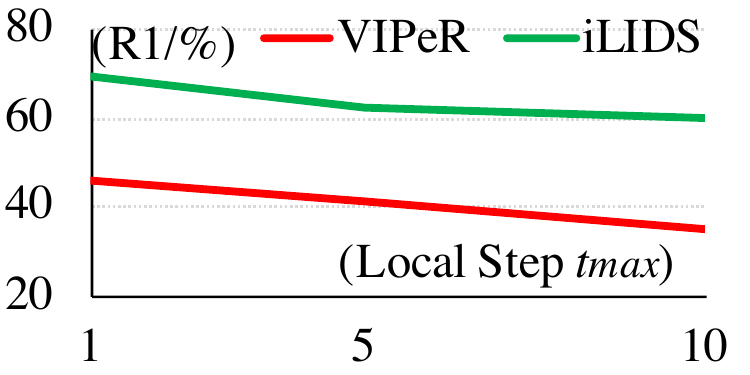}
\label{fig:local_step}
}
\subfigure[Expert regularisation]{
\includegraphics[width=0.21\linewidth]{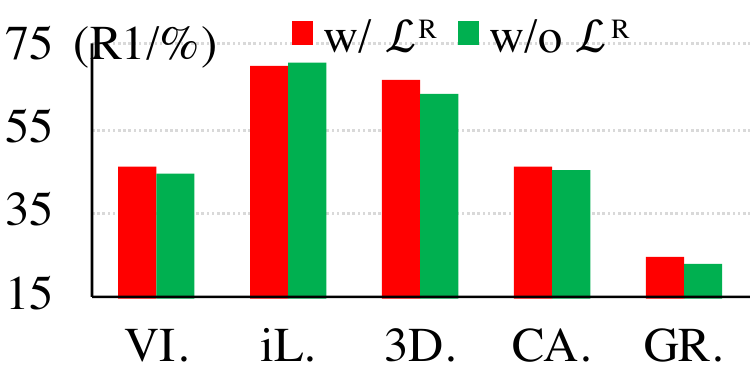}
\label{fig:expert_regularisation}
}
\caption{Ablation study on (a) client number, (b) client fraction,
(c) local steps and  (d) expert regularisation.}
\end{figure*}


{\subsection{Generalised Re-ID Performance Evaluation}}
FedReID is uniquely designed for protecting local client privacy by learning a generalisable model
without centralised sharing of training data,
whilst no existing Re-ID methods explicitly integrate privacy protection requirements into their designs.
To further show the generalisation of FedReID,
we employed out-of-the-box evaluation on non-client unseen domains without training data for fine-tuning.

{\noindent\bf Results on Smaller Benchmarks.}
We compared FedReID with:
(1) four unsupervised cross-domain fine-tuning methods
(TJAIDL~\cite{wang2018transferable}, DSTML~\cite{hu2015deep}, UMDL~\cite{Peng_2016_CVPR}, PAUL \cite{yang2019patch}),
and (2) five unsupervised generalisation methods
(SyRI~\cite{bak2018domain}, SSDAL~\cite{su2016deep},
MLDG~\cite{li2018learning},
CrossGrad~\cite{shankar2018generalizing}, DIMN~\cite{song2019generalizable}).
As shown in Table~\ref{table:state-of-the-art_small}, FedReID performs competitively against
contemporary methods,
which shows the effectiveness of the generalised global model for out-of-the-box deployment with privacy protection.
For example, 
FedReID achieves 46.5\% R1 on VIPeR and 69.7\% R1 on iLIDS, which are the second-best and close to
DIMN~\cite{song2019generalizable} which assembles all training data together without data privacy protection.
Note that, FedReID is also compatible with other techniques for better generalisation, such as style normalisation~\cite{jin2020style}.

\begin{table}[t]
\begin{center}
\small
\begin{tabular}{c|c|c|c|c}
\hline
Methods &mAP & R1 & R5 & R10\\
\hline\hline

DSIFT$^\star$+Euclidean&41.1 & 45.9 &-&-\\

BoW$^\star$+Cosine & 62.5 & 67.2 &-&-\\

DLDP$^\star$& 74.0 & 76.7 &-&-\\

\hline
FedReID & 80.4& 83.4 & 90.3& 92.4\\

Baseline (Centralised joint) & 74.7 & 77.4 & 87.2 & 90.1\\

\hline
\end{tabular}
\end{center}
\caption{Evaluation on CUHK-SYSU person search subset for Re-ID.
$^\star$: Reported results using ground-truth person images and a gallery size of 100 images per query.
}
\label{table:state-of-the-art_large}
\end{table}

{\noindent\bf Results on Large Benchmark.}
To further evaluate FedReID on a large-scale target domain,
we used the Re-ID subset of CUHK-SYSU person search dataset,
which has distinctively different scene context to most other Re-ID datasets above.
As shown in Table~\ref{table:state-of-the-art_large},
FedReID achieves competitive performance compared with some unsupervised Re-ID methods
(DSIFT~\cite{zhao2013unsupervised},
BoW~\cite{zheng2015scalable} and DLDP~\cite{schumann2017deep})
and the centralised join-training baseline,
which shows the generalisation of the global model of FedReID for
deployment on large-scale Re-ID.

\begin{table}[t]
\begin{center}
\small
\begin{tabular}{p{0.52\columnwidth}|p{0.05\columnwidth}p{0.06\columnwidth}|p{0.05\columnwidth}p{0.06\columnwidth}}
\hline
\multicolumn{1}{c|}{\multirow{2}{*}{Methods}}
&\multicolumn{2}{c|}{\multirow{1}{*}{Market}}&\multicolumn{2}{c}{\multirow{1}{*}{Duke}} \\
& R1 & mAP & R1& mAP \\

\hline\hline
Baseline(local sup.)& 88.3& 71.4 & 77.3 & 58.9 \\

FedReID(direct w/o fine-tuning)& 80.2 & 60.1 & 68.0 & 52.1 \\

FedReID(sup. w/ local data\&dis.) & 90.3 & 76.2 & 77.4 & 60.7 \\

\hline
DPR(unsup. multi-domain dis.)& 61.5 & 33.5 & 48.4 & 29.4 \\
DPR(semi-sup. multi-domain dis.)& 63.7 & 35.4 & 57.4 & 36.7 \\

\hline
\end{tabular}
\end{center}
\caption{Source labelled client tests.
   On a larger benchmark MSMT17, FedReID (direct) yields 48.4\% R1,
   FedReID (supervised) 64.6\% R1, and Baseline 61.7\% R1, respectively.
   sup.: supervised, unsup.: unsupervised, dis.: distillation.
}
\label{table:exp_source}
\end{table}

\subsection{Further Analysis and Discussion}
\label{ablation}
{\noindent\bf Source Client Tests.}
As shown in Table~\ref{table:exp_source},
although FedReID (direct) only aggregates local model updates {\emph{w/o accessing local data}},
it still achieves competitive performance,
especially when compared with DPR~\cite{wu2019distilled}
which uses local unlabelled data (unsupervised or semi-supervised).
We also tested FedReID (supervised) by using local data for model training
and a global model as the expert for distillation,
which improves FedReID (direct) and outperforms Baseline (local supervised).
Here, the improvement in R1 on Duke is small
because Duke contains many ambiguous hard negatives
and knowledge distillation helps to learn softer distribution for retrieving more positives
but is not so helpful for distinguishing hard negatives.

{\noindent\bf{Client Number $N$.}}
Fig.~\ref{fig:client_num} compares central server
aggregation with different numbers of local clients,
where $N$=1, 2 and 4 denote Market, Market+Duke and Market +Duke+Cuhk03+Msmt as clients,
respectively.
We can see that that collaboration of multi-domain clients is good
for learning more generalisable knowledge in the central server.

{\noindent\bf{Client Fraction $S$.}}
Fig.~\ref{fig:client_frac} compares FedReID with different client fractions
$S$.
We can see that updating with an arbitrary client ($S$=0.25) is inferior to aggregating
multiple clients,
whilst aggregating all clients ($S$=1.0) performs slightly better than
aggregating randomly selected clients ($S$=0.5),
but random selection protects more data privacy.

{\noindent\bf{Client Local Step $t_{max}$.}}
Fig.~\ref{fig:local_step} compares FedReID with different client local steps
which potentially promote communication efficiency.
Overall, the performance of
FedReID gradually decreases when $t_{max}$ increases due to the accumulation of biases in each local client.

{\noindent\bf Local Expert Regularisation}.
Fig.~\ref{fig:expert_regularisation} shows the evaluation on the
regularisation of local experts.
With the expert regularisation,
FedReID gets better generalisation overall,
which shows that the expert regularisation provides richer knowledge
to facilitate the improvement of FedReID.

\section{Conclusion}
In this work, we introduced decentralised learning from
independent multi-domain label spaces for person Re-ID
and proposed a new paradigm called Federated Person Re-Identification (FedReID).
We trained multiple local models at different clients
using non-shared private local data
and aggregated local model updates to construct a global model in a central server.
This iterative client-server collaborative learning helps to build a generalisable model for deployment.
Extensive experiments on ten Re-ID datasets show the unique advantage of FedReID
over contemporary methods.

\section{Acknowledgements}
This work is supported by Vision Semantics Limited,
Alan Turing Institute Turing Fellowship, and Innovate UK Industrial Challenge Project on Developing and Commercialising
Intelligent Video Analytics Solutions for Public Safety (98111-571149), Queen Mary University of London Principal’s Scholarship.

\bibliography{WuEtAl_FedReID}
\end{document}